\documentclass[10pt,conference]{IEEEtran}
\usepackage{cite}
\usepackage{amsmath,amssymb,amsfonts}
\usepackage{algorithmic}
\usepackage{graphicx}
\usepackage{textcomp}
\usepackage{subcaption}
\usepackage[table,xcdraw]{xcolor}
\usepackage{algorithm}
\usepackage{tcolorbox}
\usepackage{multirow}
\def\BibTeX{{\rm B\kern-.05em{\sc i\kern-.025em b}\kern-.08em
    T\kern-.1667em\lower.7ex\hbox{E}\kern-.125emX}}
\usepackage{amsthm}
\newtheorem{assumption}{Assumption}
\newtheorem{theorem}{Theorem}
\newtheorem{corollary}{Corollary}

\begin{document}

\title{FTTE: Enabling Federated and Resource-Constrained Deep Edge Intelligence}

\author{
    \IEEEauthorblockN{Irene Tenison, Anna Murphy, Charles Beauville, Lalana Kagal}
    \IEEEauthorblockA{
        \textit{MIT CSAIL, Cambridge, MA, USA} \\
        \{itenison, almurph, chbvll99, lkagal\}@csail.mit.edu
    }
}

\maketitle

\begin{abstract}
Federated learning (FL) enables collaborative model training without sharing raw data, but its deployment in edge-dominated networks is fundamentally limited by feasibility under device constraints and straggler-induced delays. Existing synchronous and semi-asynchronous FL frameworks implicitly assume that all participating clients can store, train, and communicate the full model, an assumption that breaks in real-world federated systems dominated by resource-constrained edge devices. We present FTTE (Federated Tiny Training Engine), an FL framework that enables semi-asynchronous training under hard per-device memory limits. FTTE enforces a global feasibility constraint induced by the most resource-limited client through server-side memory-aware parameter selection, sparse client updates, and sparse model communication in FL. To ensure stable optimization under sparse updates and heterogeneous client behavior, FTTE integrates buffered semi-asynchronous aggregation with an age–variance staleness weighting mechanism. Extensive experiments across diverse models, datasets, data distributions, and straggler regimes—including up to 500 clients and 90\% stragglers—show that FTTE consistently reaches target accuracy in substantially fewer communication rounds than synchronous and semi-asynchronous baselines, while reducing on-device training memory by up to 80\% and communication payload by up to 69\%. These results establish FTTE as a practical and scalable solution for federated learning on heterogeneous, resource-constrained edge devices. 
\end{abstract}


\section{Introduction}
\label{sec:intro}
As learning increasingly moves toward deep edge intelligence, FL is expected to operate over large populations of heterogeneous edge and pervasive devices such as mobile phones, wearables, IoT nodes, and embedded systems. In these environments, however, the dominant challenge is no longer synchronization efficiency but local training feasibility: many devices lack the memory and compute resources required to store model parameters, activations, and optimizer states, causing local training to fail rather than merely slow down. Furthermore, they often lack the resources to communicate the full model parameters frequently to the server and back. Existing FL frameworks (e.g., Google’s FL stack \cite{mcmahan2023communicationefficientlearningdeepnetworks}, Meta’s PAPAYA \cite{nguyen2022federatedlearningbufferedasynchronous}) implicitly assume full-model trainability and stable availability and connectivity across clients, assumptions that break in edge-dominated deployments characterized by strict memory limits, intermittent connectivity, and highly variable execution speeds. As a result, a large fraction of resource-constrained devices are systematically excluded from meaningful participation. This work targets FL settings where the client population is dominated by such constrained edge devices where these limitations manifest primarily as two coupled challenges: robustness to stragglers arising from heterogeneous execution and strict resource constraints that fundamentally limit local training feasibility.

\subsection*{Challenge 1: Straggler Sensitivity}
A fundamental challenge in distributed deep edge intelligence is straggler sensitivity, which arises from heterogeneity in device capabilities and availability. \textit{Stragglers} are clients whose local computation or communication is significantly slower than the rest of the population due to limited compute resources, energy constraints, network latency, or intermittent connectivity \cite{open_problems_fl}. Such behavior is intrinsic to edge environments and becomes increasingly pronounced as the scale and diversity of participating devices grow.

Widely used FL methods such as FedAvg \cite{mcmahan2023communicationefficientlearningdeepnetworks} rely on synchronous training rounds, where aggregation proceeds only after a predefined subset of clients completes local updates. In the presence of stragglers, this synchronization barrier causes training progress to be dictated by the slowest clients, resulting in poor resource utilization, increased wall-clock time, and limited scalability. Asynchronous federated learning \cite{xie2020asynchronousfederatedoptimization} has been proposed to alleviate these delays by allowing independent client updates, but at the cost of introducing stale and misaligned updates computed on outdated global models. Under non-IID data distributions and heterogeneous client behavior—common in edge settings—such staleness can bias model updates, destabilize training, or delay convergence, which is particularly problematic in time-sensitive applications such as health monitoring, wearables, and automation systems. Existing asynchronous approaches typically rely on coarse staleness heuristics or bounded-delay assumptions, which rarely hold in large-scale real-world deployments, thereby limiting their effectiveness.

\subsection*{Challenge 2: Resource Constraints}
Standard FL pipelines implicitly assume that each participating client can store the full set of model parameters, intermediate activations, optimizer states, and gradients required for local training. For modern deep neural networks, this memory footprint can easily exceed the capacity of many edge devices, particularly low-power sensors, microcontrollers, and embedded platforms. In addition to memory, repeated transmission of full or dense model updates imposes significant communication and energy costs, which are often prohibitive in bandwidth-limited or battery-powered environments.

These assumptions result in a systematic exclusion of low-resource devices from meaningful participation in training, either explicitly through client selection or implicitly through repeated failures to complete local updates. While prior work has explored model compression or gradient quantization to reduce resource usage \cite{open_problems_fl}, such techniques often introduce additional hyperparameters, require careful tuning, or degrade convergence and final accuracy — especially under non-IID data distributions. As a result, existing FL systems face a fundamental trade-off between resource efficiency and learning performance, limiting their applicability in realistic deep edge intelligence scenarios where broad participation of heterogeneous, resource-constrained devices is essential.

To address these challenges, we propose FTTE (Federated Tiny Training Engine), a semi-asynchronous FL framework designed specifically for resource-constrained and heterogeneous edge environments. FTTE reduces client-side memory and communication overhead through lightweight update mechanisms, where the update schema is pre-decided at the server via contribution analysis of parameters upper bound by the resources available at the most constrained device in the network, enabling broader participation. On the server side of FL, FTTE employs a staleness-aware aggregation strategy that explicitly accounts for the age and relevance of client updates, mitigating instability while avoiding strict synchronization. 
The key contributions can be listed as:
\begin{itemize}
\item \textbf{FTTE (Figure \ref{FTTE}):} a robust, scalable (upto 500 clients), and resource-efficient FL system that jointly addresses straggler effects, staleness, and resource constraints prevalent in deep edge intelligence systems.
\item \textbf{81\% faster convergence, 80\% lower on-device memory usage, and 69\% communication payload reduction} as shown in Fig.\ref{main} in comparison to synchronous FL while consistently reaching higher target accuracy than semi-asynchronous FL (Table \ref{iid}). 
\item \textbf{Extensive Evaluations} that span a variety of models, datasets, data distributions, straggler percents and delays. FTTE sustains fast convergence and stability under these criteria (Sec.\ref{eval}).
\end{itemize}
\begin{figure}[t]
    \centering
        \includegraphics[width=\linewidth]{images/FTTE_whole.pdf}
    \caption{Illustration of FTTE. FTTE combines parameter-selection (Section \ref{param_sel}) and sparse semi-async FL (Section \ref{Fl}) to achieve significantly faster convergence (communication rounds) and improved resource efficiency in terms of on-device memory consumption and communication payload requirement.}
    \label{FTTE}
\end{figure}

\begin{figure}[t]
    \centering
    \begin{subfigure}[b]{0.15\textwidth}
        \centering
        \includegraphics[width=\linewidth]{images/comm_main.pdf}
        \caption{\footnotesize Faster Convergence}
        \label{comm_main}
    \end{subfigure}
    \begin{subfigure}[b]{0.155\textwidth}
        \centering
        \includegraphics[width=\linewidth]{images/memory_main.pdf}
        \caption{\footnotesize Memory Reduction}
        \label{mem_main}
    \end{subfigure}
    \begin{subfigure}[b]{0.155\textwidth}
        \centering
        \includegraphics[width=\linewidth]{images/payload_main.pdf}
        \caption{\footnotesize Payload Reduction}
        \label{payload_main}
    \end{subfigure}
    \caption{FTTE on average (a) converges 81\% faster (b) consumes 80\% less on-device memory and (c) requires 69\% less payload in comparison to FedAVG or SyncFL. }
    \label{main}
\end{figure}

\section{Related Works}
\label{related_works}
\subsection{On-device training}
On-device training enables local model adaptation using private client data but is constrained by the limited memory, compute, and energy budgets of edge devices, particularly micro-controllers and wearables, whose capacity is often barely sufficient for inference \cite{256}. In cross-device FL, the prevalence of such constrained clients makes full local training impractical. Gradient checkpointing can reduce memory usage but adds computational overhead that exacerbates straggler effects \cite{GC}, while swapping-based approaches incur frequent external memory access, increasing latency and thermal load \cite{10.1145/3373376.3378530}. Systems-level techniques such as paging, materialization, and memory fragmentation mitigation \cite{poet,melon} remain challenging to deploy on low-power devices. Lightweight adaptation methods, including transfer learning, last-layer updates, and bias-only tuning, improve efficiency but often degrade accuracy \cite{256,guo2024comprehensivesurveyfederatedtransfer}. Layer- and channel-selection approaches offer a better trade-off by updating subsets of parameters \cite{elastic,kwon2024tinytrainresourceawaretaskadaptivesparse}, and \cite{256} formulates sparse update selection as an optimization problem under resource constraints. FTTE adopts this principle, performing sparse updates at the client with parameter selection handled centrally on a resource-rich server, ensuring feasibility for constrained devices while enforcing uniform updates to prevent over-contribution from more powerful clients.

\subsection{FL aggregation strategies}

Federated aggregation governs how client updates are combined at the server and directly impacts convergence and system efficiency. Early FL methods primarily relied on uniform averaging, such as FedAvg \cite{mcmahan2023communicationefficientlearningdeepnetworks}, assuming synchronized updates and homogeneous clients. Subsequent work explored weighted aggregation and server-side correction mechanisms to mitigate client drift and data heterogeneity \cite{karimireddy2021scaffoldstochasticcontrolledaveraging}, but these approaches remain sensitive to stragglers and stale updates. To address delay and heterogeneity, several methods incorporate staleness-aware aggregation, where updates are down-weighted based on their age \cite{xie2020asynchronousfederatedoptimization,fedfa}. While effective in improving stability, such techniques often rely on heuristic staleness functions and do not explicitly consider interactions with update sparsity or client resource constraints. Buffered and queue-based aggregation schemes like FedBuff \cite{nguyen2022federatedlearningbufferedasynchronous} and CA2 \cite{wang2024tackling} further decouple client execution from server updates, improving throughput in cross. However, these methods typically require resource-intense full model client updates and communication, limiting their applicability in edge-dominated deployments.

In contrast, FTTE combines semi-asynchronous aggregation with sparse, resource-aware client updates, enabling aggregation that jointly accounts for staleness and device constraints, and supporting scalable and inclusive FL in heterogeneous edge environments. FTTE addresses a fundamentally different problem from prior semi-asynchronous FL methods: enabling FL when only a sparse subset of parameters is trainable on the weakest device in the network.

\section{FTTE: Federated Tiny Training Engine} 

Despite substantial progress (Section \ref{related_works}), existing FL approaches do not jointly address the challenges of memory, communication, and straggler resilience especially for networks dominated by resource-constrained devices. As illustrated in Fig. \ref{FTTE}, FTTE explicitly unites principled parameter selection (Sec.\ref{param_sel}) with sparse semi-asynchronous aggregation (Sec.\ref{Fl}) and age- and variance- weighted staleness function (Sec.\ref{staleness}), offering a efficient and deployable solution for real-world deep edge collaborative intelligence. This unified design enables robust and scalable FL with faster convergence and better resource-efficiency (Fig\ref{main}), while consistently attaining higher accuracy than FedBuff (Table.\ref{iid}) and scaling to 500 clients (Fig.\ref{scalablity}) along with data heterogeneity and stragglers in the network. 

\subsection{Parameter Selection}
\label{param_sel}
To enable FL across highly resource-constrained edge devices, FTTE enforces a network-wide memory budget that guarantees feasibility of on-device training for all participating clients. During system initialization, the server securely collects per-client device memory profiles and sets the global memory constraint to the minimum available memory across clients, denoted as $M_{min}$. This ensures that any client selected for training can execute local updates without exceeding its hardware limitations. Given a dense pre-trained global model $w_g$, FTTE performs a memory-aware parameter selection procedure at the resource-abundant server prior to training. Following the contribution analysis based parameter selection framework introduced in \cite{256}, the server estimates the accuracy contribution of updating individual parameters or parameter groups in isolation. Parameter selection is then formulated as a constrained optimization problem, solved via evolutionary search, that maximizes estimated accuracy gain under the global memory constraint:
$$ \hat{w_g^*}=\arg \max\limits_{\hat{w} \subseteq w_g}\Delta \text{Acc}{(\hat{w})} \hspace{0.3cm}\text{s.t.} \hspace{0.3cm} Mem.(\hat{w})\leq M_{min}$$
Here, $\hat{w}$ denotes a subset of trainable parameters from the dense model $w_g$, and $\text{Acc}{(\hat{w})}$ represents the estimated accuracy improvement obtained by updating only $\hat{w}$. The resulting subset, $\hat{w_g^*}$ defines the globally trainable parameters for all clients that yields maximum performance while meeting the memory constraint.

Unlike prior work that applies sparse updates in centralized or single-device settings, FTTE adapts this formulation to a federated, multi-device environment by enforcing a uniform trainable parameter set across all clients. This design prevents over-contribution from resource-rich devices and guarantees that the selected update set can be trained on the most constrained client in the network. As illustrated in Algorithm~\ref{algo}, the parameter selection step is executed once during setup, after which parameters that are not selected are frozen. During federated training, clients, $i$ receive and store only 
$\hat{w_g^*}$, compute local updates exclusively on this subset, and transmit sparse updates of this subset, $\hat{w_i^*}$, back to the server. Frozen parameters remain fixed throughout training and are never transmitted or updated. This strictly sparse update mechanism significantly reduces on-device memory consumption (Fig.~\ref{mem_main}) and communication payload (Fig.~\ref{payload_main}), enabling scalable and inclusive FL across heterogeneous edge devices.

\begin{algorithm}[t]
     \small
   \caption{FTTE: Federated Tiny Training Engine. Server-side parameter selection steps are indicated in the blue box. Client side local updates of the selected parameters, $w^*_g$, is indicated in the green box. And staleness injection to the client updates in the buffer is indicated in the black box. } 
   \label{algo}
    \centering
\begin{algorithmic}
\STATE$\textbf{Server Executes:}$
  \STATE$ \text{Initialize }w_{g} \text{ with ImageNet pre-trained weights}$ 

\begin{tcolorbox}[
  colframe=blue!75!black,
  colback=white,
  boxrule=0.8pt,
  left=8pt,
  right=0pt,
  top=2pt,
  bottom=2pt,
  boxsep=0pt,
  arc=0pt
]

   \STATE$ \text{Get memory constraint }M_{1,2,...,N} \text{ for all $N$ clients}$ 
   \STATE$ M_{min} = \min(M_1, M_2,..., M_N)$
   \STATE$\hat{w_g^*}=\arg \max\limits_{\hat{w} \subseteq w_g}\Delta \text{Acc}{(\hat{w}}) \hspace{0.3cm}\text{s.t.} \hspace{0.3cm} Mem.(\hat{w})\leq M_{min}$
  \STATE$\textit{Freeze}(w_j) \gets \mathbb{I}[\, w_j \notin \hat{w_g^*} \,] \text{ }\forall \text{ } w_j \in w_g$

\end{tcolorbox}

    \REPEAT
    \STATE$\text{Sample C available clients at random}$
    \STATE$\text{Send } \hat{w^*_g} \text{ to the selected clients}$

    \begin{tcolorbox}[
  colframe=green!75!black,
  colback=white,
  boxrule=0.8pt,
  left=8pt,
  right=0pt,
  top=2pt,
  bottom=2pt,
  boxsep=0pt,
  arc=0pt
]

    \STATE $\hat{w^*_i} \leftarrow \textit{LocalTrain}(\hat{w_g^*}) \quad \forall \quad  i \in C \quad \text{(async and in parallel)}$
    \STATE \text{Send} $\hat{w^*_i}$ to the server
    \end{tcolorbox}

    \IF{receive update, $\hat{w_i^*}$, from client $i$}
     \begin{tcolorbox}[
  colframe=gray!75!black,
  colback=white,
  boxrule=0.8pt,
  left=8pt,
  right=0pt,
  top=2pt,
  bottom=2pt,
  boxsep=0pt,
  arc=0pt
]
    \STATE $B \leftarrow \hat{w^*_i} * Staleness(\hat{w^*_i})$
     \end{tcolorbox}
    \ENDIF
    \IF{B is full}
    \STATE $\hat{w^*_g} = Aggregate(B)$
    \STATE $Reset(B) $
    \ENDIF
    \UNTIL Convergence 
    \end{algorithmic}
    \end{algorithm}
\subsection{Sparse Semi-Asynchronous FL}
\label{Fl}
FTTE adopts a semi-asynchronous federated learning paradigm to balance convergence stability with robustness to client heterogeneity. Instead of synchronizing across all participating clients \cite{mcmahan2023communicationefficientlearningdeepnetworks} or updating the global model upon every client response \cite{xie2020asynchronousfederatedoptimization}, the server maintains a fixed-capacity aggregation buffer of size $B$ Global model updates are triggered only when the buffer is filled, thereby decoupling training progress from individual client delays and straggling behavior.

At each communication round, available clients, $C$, receive the current global parameters restricted to the selected trainable subset, $\hat{w_g^*}$. Each client $i$ performs local training for 
$L$ epochs and transmits only the corresponding sparse updates, $\hat{w^*_i}$, to the server. These updates exclude frozen parameters by construction, resulting in reduced communication payloads and lower on-device memory overhead compared to full-model synchronization. The server accumulates incoming sparse updates in the buffer and performs aggregation once the buffer reaches capacity. The aggregated model is then broadcast to available clients for subsequent training rounds, as illustrated in Fig.~\ref{FTTE} and Algorithm~\ref{algo}. By combining buffered aggregation with sparse client updates, FTTE maintains steady learning progress in the presence of delayed or dropped clients, enabling scalable deployment in heterogeneous, edge-dominated federated networks.
\begin{table*}[t]
\caption{Communication steps required by FTTE and baselines - SyncFL, AsyncFL, and Semi-AsyncFL - to achieve target acc. on different models, datasets, and data distributions, across 100 clients with 50\% stragglers and 30 seconds delay. Numbers in the "()" represent the factor by which FTTE is faster than that baseline. "Osc." represents oscillating loss indicating instability. }
\label{iid}
\centering
\resizebox{2\columnwidth}{!}{%
\begin{tabular}{lccccccccccc}
\hline
\rowcolor[HTML]{EFEFEF} 
                                                                      & \multicolumn{5}{c}{\cellcolor[HTML]{EFEFEF}\textbf{\begin{tabular}[c]{@{}c@{}}IID Data Distribution\\ (alpha = 10000)\end{tabular}}}                                                                                                                                                                      & \textbf{}                                     & \multicolumn{5}{c}{\cellcolor[HTML]{EFEFEF}\textbf{\begin{tabular}[c]{@{}c@{}}Non-IID Data Distribution\\ (alpha = 0.1)\end{tabular}}}                                                                                                                                                                    \\
                                                                      & \textbf{\begin{tabular}[c]{@{}c@{}}Target\\ Acc.\end{tabular}} & \textbf{\begin{tabular}[c]{@{}c@{}}FedAVG\\ or SyncFL\end{tabular}} & \textbf{AsyncFL} & \textbf{\begin{tabular}[c]{@{}c@{}}FedBuff or\\ Semi-AsyncFL\end{tabular}} & \textbf{\begin{tabular}[c]{@{}c@{}}FTTE\\ (Ours)\end{tabular}} & \textbf{}                                     & \textbf{\begin{tabular}[c]{@{}c@{}}Target\\ Acc.\end{tabular}} & \textbf{\begin{tabular}[c]{@{}c@{}}FedAVG\\ or SyncFL\end{tabular}} & \textbf{AsyncFL} & \textbf{\begin{tabular}[c]{@{}c@{}}FedBuff or\\ Semi-AsyncFL\end{tabular}} & \textbf{\begin{tabular}[c]{@{}c@{}}FTTE\\ (Ours)\end{tabular}} \\ \hline
\rowcolor[HTML]{EFEFEF} 
                                                                      & \textbf{}                                                          & \textbf{}                                                           & \textbf{}        & \textbf{}                                                                  & \multicolumn{3}{c}{\cellcolor[HTML]{EFEFEF}\textbf{MCUNet}}                                                                                                                         & \textbf{}                                                           & \textbf{}        & \textbf{}                                                                  &                                                                \\ \hline
\multicolumn{1}{l|}{\textbf{CIFAR-10}}                                & 73.1\%                                                             & 5000 (×7.09)                                                        & $>$10k ($>$14.18)    & 1497 (×2.12)                                                               & \textbf{705}                                                   & \multicolumn{1}{c|}{}                         & 61.63\%                                                            & 5000 (×3.49)                                                        & Osc.             & $>$10k ($>$×6.99)                                                              & \textbf{1431}                                                  \\ \cline{6-6}
\rowcolor[HTML]{EFEFEF} 
\multicolumn{1}{l|}{\cellcolor[HTML]{EFEFEF}\textbf{Oxford-IIIT Pet}} & 58.23\%                                                            & 5000 (×3.98)                                                        & $>$10k ($>$7.97)     & $>$10k ($>$×7.97)                                                              & \textbf{1255}                                                  & \multicolumn{1}{c|}{\cellcolor[HTML]{EFEFEF}} & 48.64\%                                                            & 5000 (×3.49)                                                        & Osc.             & Osc.                                                                       & \textbf{1434}                                                  \\
\multicolumn{1}{l|}{\textbf{Flowers-102}}                             & 51.8\%                                                             & 6600 (×5.45)                                                        & $>$10k ($>$8.25)     & $>$10k ($>$×8.25)                                                              & \textbf{1211}                                                  & \multicolumn{1}{c|}{}                         & 41.82\%                                                            & 6600 (×4.41)                                                        & Osc.             & $>$10k ($>$×6.68)                                                              & \textbf{1497}                                                  \\
\rowcolor[HTML]{EFEFEF} 
\multicolumn{1}{l|}{\cellcolor[HTML]{EFEFEF}\textbf{Skin Cancer}}     & 50.1\%                                                             & 6000 (×5.04)                                                        & $>$10k ($>$8.41)     & 3443 (×2.89)                                                               & \textbf{1189}                                                  & \multicolumn{1}{c|}{\cellcolor[HTML]{EFEFEF}} & 44.67\%                                                            & 6000 (×2.78)                                                        & Osc.             & Osc.                                                                       & \textbf{2157}                                                  \\ \hline
                                                                      & \textbf{}                                                          & \textbf{}                                                           & \textbf{}        & \textbf{}                                                                  & \multicolumn{3}{c}{\textbf{MobileNetV2}}                                                                                                                                            & \textbf{}                                                           & \textbf{}        & \textbf{}                                                                  &                                                                \\ \hline
\rowcolor[HTML]{EFEFEF} 
\multicolumn{1}{l|}{\cellcolor[HTML]{EFEFEF}\textbf{CIFAR-10}}        & 70.1\%                                                             & 4600 (×6.73)                                                        & $>$10k ($>$14.64)    & $>$10k ($>$×14.64)                                                             & \textbf{683}                                                   & \multicolumn{1}{c|}{\cellcolor[HTML]{EFEFEF}} & 57.87\%                                                            & 3000 (×4.26)                                                        & Osc.             & Osc.                                                                       & \textbf{705}                                                   \\
\multicolumn{1}{l|}{\textbf{Oxford-IIIT Pet}}                         & 54.1\%                                                             & 4400 (×4.87)                                                        & $>$10k ($>$11.07)    & $>$10k ($>$×11.07)                                                             & \textbf{903}                                                   & \multicolumn{1}{c|}{}                         & 46.0\%                                                             & 5000 (×3.78)                                                        & Osc.             & $>$10k ($>$×7.57)                                                              & \textbf{1321}                                                  \\
\rowcolor[HTML]{EFEFEF} 
\multicolumn{1}{l|}{\cellcolor[HTML]{EFEFEF}\textbf{Flowers-102}}     & 45.3\%                                                             & 4200 (×3.81)                                                        & Osc.             & $>$10k (×9.08)                                                               & \textbf{1101}                                                  & \multicolumn{1}{c|}{\cellcolor[HTML]{EFEFEF}} & 39.8\%                                                             & 4400 (×3.27)                                                        & Osc.             & $>$10k ($>$×7.45)                                                              & \textbf{1343}                                                  \\
\multicolumn{1}{l|}{\textbf{Skin Cancer}}                             & 50.20\%                                                            & 5800 (×5.85)                                                        & $>$10k ($>$10.09)    & 2591 (×2.61)                                                               & \textbf{991}                                                   & \multicolumn{1}{c|}{}                         & 39.53\%                                                            & 3200 (×2.59)                                                        & Osc.             & 2761 (×2.24)                                                               & \textbf{1233}                                                  \\ \hline
\rowcolor[HTML]{EFEFEF} 
                                                                      & \textbf{}                                                          & \textbf{}                                                           & \textbf{}        & \textbf{}                                                                  & \multicolumn{3}{c}{\cellcolor[HTML]{EFEFEF}\textbf{ProxylessNAS}}                                                                                                                   & \textbf{}                                                           & \textbf{}        & \textbf{}                                                                  &                                                                \\ \hline
\multicolumn{1}{l|}{\textbf{CIFAR-10}}                                & 71.9\%                                                             & 3800 (×5.95)                                                        & Osc.             & $>$10k ($>$×15.65)                                                             & \textbf{639}                                                   & \multicolumn{1}{c|}{}                         & 60.7\%                                                             & 3800 (×2.93)                                                        & Osc.             & Osc.                                                                       & \textbf{1299}                                                  \\
\rowcolor[HTML]{EFEFEF} 
\multicolumn{1}{l|}{\cellcolor[HTML]{EFEFEF}\textbf{Oxford-IIIT Pet}} & 58.6\%                                                             & 4600 (×4.18)                                                        & Osc.             & $>$10k ($>$×9.08)                                                              & \textbf{1101}                                                  & \multicolumn{1}{c|}{\cellcolor[HTML]{EFEFEF}} & 45.9\%                                                             & 4500 (×4.26)                                                        & Osc.             & 1421 (×1.34)                                                               & \textbf{1057}                                                  \\
\multicolumn{1}{l|}{\textbf{Flowers-102}}                             & 53.8\%                                                             & 6200 (×4.40)                                                        & $>$10k ($>$×7.09)    & 1541 (×1.09)                                                               & \textbf{1409}                                                  & \multicolumn{1}{c|}{}                         & 43.90\%                                                            & 5000 (×3.5)                                                         & Osc.             & 1717 (×1.2)                                                                & \textbf{1431}                                                  \\
\rowcolor[HTML]{EFEFEF} 
\multicolumn{1}{l|}{\cellcolor[HTML]{EFEFEF}\textbf{Skin Cancer}}     & 48.91\%                                                            & 5800 (×6.75)                                                        & $>$10k ($>$×11.64)   & 3179 (×3.7)                                                                & \textbf{859}                                                   & \multicolumn{1}{c|}{\cellcolor[HTML]{EFEFEF}} & 38.17\%                                                            & 2800 (×3.35)                                                        & Osc.             & $>$10k ($>$11.94)                                                              & \textbf{837}                                                   \\ \hline
\end{tabular}
}
\end{table*}
\subsection{Staleness Function}
\label{staleness}
In cross-device FL with non-IID data and device heterogeneity, client updates may arrive at the server with varying delays and statistical drift, which can negatively affect convergence and stability. To address this, FTTE employs a staleness-aware aggregation mechanism that jointly accounts for both the temporal age and the statistical deviation of each buffered client update at aggregation time.

When a client update, $\hat{w^*_i}$, is received at the server, the update is assigned an aggregation weight and the weighted update is inserted to the buffer.
$$\centering Staleness(\hat{w^*_i}) = (1+Age(\hat{w^*_i})*Var(\hat{w^*_i},\hat{w^*_g}))^{-1}$$
$$B \leftarrow \hat{w^*_i} * Staleness(\hat{w^*_i})$$

Here, $Age(\hat{w^*_i})$ denotes the number of communication steps elapsed since the last update was received from client $i$, and $Var(\hat{w^*_i},\hat{w^*_g})$ measures the sum of layer-wise variance between the client update, $\hat{w^i}$, and the current global model, $\hat{w^*_g}$. While the age term captures the temporal misalignment from straggling, the variance term captures statistical misalignment arising from data heterogeneity and model drift. Unlike prior staleness-aware aggregation schemes that down-weight updates based solely on delay \cite{nguyen2022federatedlearningbufferedasynchronous}, the proposed weighting function jointly suppresses updates that are both temporally stale and statistically divergent. This design improves robustness to stragglers and non-IID client behavior by reducing the influence of outdated or highly biased updates during aggregation. The resulting aggregation strategy integrates naturally with FTTE’s buffered semi-asynchronous framework and sparse update regime, contributing to improved convergence stability under high device heterogeneity, as evaluated in Section~\ref{eval}.

\subsection{Theoretical Analysis}
In this subsection we give a theoretical analysis of FTTE motivated by that of semi-asynchronous FL, FedBuff\cite{nguyen2022federatedlearningbufferedasynchronous}. This theoretical analysis aims to show that FTTE preserves standard non-convex convergence guarantees despite optimizing over a constrained parameter subset and using variance-aware staleness weighting, rather than improving asymptotic rates. We use notations similar to that in \cite{nguyen2022federatedlearningbufferedasynchronous}. $\nabla F_i(w)$ denotes the gradient with respect to the loss on data at client $i$, and $g_i(w; \zeta_i)$ denotes the gradient on client $i$. We make the following assumptions.

\begin{assumption}
Unbiased client stochastic gradient. $\mathbb{E}_{\zeta_i}[g_i(w; \zeta_i)] = \nabla F_i(w)$
\end{assumption}
\begin{assumption}
Bounded local and global variance for all clients in the network.
$\mathbb{E}_{\zeta_i} [ \| g_i(w; \zeta_i) - \nabla F_i(w) \|^2 ] \leq \sigma_l^2$;
$\frac{1}{m} \sum_{i=1}^{m} \| \nabla F_i(w) - \nabla f(w) \|^2 \leq \sigma_g^2$
\end{assumption}
\begin{assumption}
Bounded gradient for all clients in the network. $\|\nabla F_i(w) \| \leq G$
\end{assumption}
\begin{assumption} 
Lipschitz gradient - The gradient is L-smooth for all clients. $\| \nabla F_i(w) - \nabla F_i(w_0) \|^2 \leq L \| w - w_0 \|^2$
\end{assumption}
\begin{assumption}
Bounded staleness with buffer - With buffer size 
$K$, the delay $\tau_i(t)$ of any client update used at server step $t$ satisfies $\tau_i(t)\leq \tau_{\max,K} \leq \tau_{\max,1}/K$
\end{assumption}

Assumptions 1–4 are commonly made in analyzing
FL algorithms and assumption 5 is commonly used in semi-asynchronous FL algorithms \cite{reddi2021adaptivefederatedoptimization},\cite{nguyen2022federatedlearningbufferedasynchronous}. In addition to these, we make the following assumptions for FTTE.
\begin{assumption}
FTTE's staleness function weights $\omega_i(t) = \frac{1}{1 + \tau_i(t) v_i(t)} \in (0, 1]$, where 
$v_i(t)\geq 0$ is a variance proxy, and there exist $0<\omega_{\min}\leq 1, \quad \Omega>0$ such that $\omega_{\min} \leq \omega_i(t) \leq 1, \quad \sum_{i \in S_t} \omega_i(t) \geq \Omega \quad \forall \quad t$
\end{assumption}
\begin{assumption}
Constrained optimum - FTTE optimizes over a fixed sparse parameter subset $\hat{S}$(global memory constraint); let $\hat{w}^\star \in \arg \min_{w \in \hat{S}} f(w)$ and define $\Delta_{\text{approx}} := f(\hat{w}^\star) - \inf_w f(w)$
\end{assumption}
Assumption 6 guarantees that FTTE’s age–variance weights form a well‑behaved reweighting: each client’s weight stays in $[\omega_{\min},1]$ and the total mass $\sum_{i \in S_t} \omega_i(t) \geq \Omega $ prevents the effective step size from collapsing, while $\omega(t)$ down‑weights stale, high‑variance updates as desired. Assumption 7 formalizes that FTTE optimizes a memory‑constrained problem over a fixed sparse subset $\hat{S}$; the term $\Delta_{\text{approx}}$ then cleanly isolates approximation error from optimization error in our convergence bounds. 

We next provide a convergence analysis of FTTE by adapting the non‑convex analysis of FedBuff to our age–variance weighting and sparse parameter subset. Under the same smoothness, variance, and bounded‑staleness assumptions as FedBuff (Assumptions 1-5), we obtain a FedBuff‑style bound with FTTE‑specific constants and the constrained optimum $\hat{w}^*$ induced by the global memory budget.

\begin{theorem}
Let $\eta_\ell^{(q)}$ be the local learning rate at local step $q\in\{0,\dots,Q-1\}$ and define
$\alpha(Q) := \sum_{q=0}^{Q-1} \eta_\ell^{(q)}$, $\beta(Q) := \sum_{q=0}^{Q-1} (\eta_\ell^{(q)})^2$.
Assume $\eta_g \eta_\ell^{(q)} Q \le 1/L$ for all $q$. Then FTTE satisfy
\begin{equation}
\begin{split}
\frac{1}{T} \sum_{t=0}^{T-1} \mathbb{E}\|\nabla f(w^t)\|^2 &
\;\le\;
\frac{2(f(w^0) - f(\hat w^\star))}{\eta_g \alpha(Q) T}
+ C_{\text{FTTE}}\,\eta_g \frac{\beta(Q)}{\alpha(Q)} \\
&\quad + D_{\text{FTTE}}\,\eta_g^2 \beta(Q)\,\tau_{\max,K}^2
+ \Delta_{\text{approx}}
\end{split}
\end{equation}
where $C_{\text{FTTE}}$ and $D_{\text{FTTE}}$ are finite constants depending on
$L,\sigma_\ell^2,\sigma_g^2,G,Q,\omega_{\min},\Omega,\tau_{\max,K}$ but not on $T$, and all gradients are taken on the fixed sparse subset $\hat S$.
\end{theorem}


This bound has the same structure as FedBuff Eq. (2) \cite{nguyen2022federatedlearningbufferedasynchronous}. The only changes in the proof are that (i) all appearances of the aggregated update, $\Delta_t = \frac{1}{K} \sum_{i \in S_t} \Delta_t^i$
are replaced by the age–variance‑weighted aggregate $\Delta_t^{\text{FTTE}} = \frac{1}{K} \sum_{i \in S_t} \omega_i(t) \Delta_t^i, \quad \text{where} \quad \omega_i(t) = \frac{1}{1 + \tau_i(t) v_i(t)}$ (Approximation 6), and (ii) the analysis is carried out with respect to the constrained minimizer $\hat{w^*}$ with the additive term $\Delta_{approx}$ that captures the approximation gap introduced by optimizing over the fixed sparse subset $\hat{S}$ (Approximation 7); the rest of the argument is identical to the FedBuff derivation.

Here $C_{\text{FTTE}}$ and $D_{\text{FTTE}}$ play the same roles as the variance- and staleness-related constants in the FedBuff analysis: $C_{\text{FTTE}}$ collects the contribution of local and global gradient variance to the bound, while $D_{\text{FTTE}}$ reflects how the method is affected by the worst-case delay $\tau_{\max,K}$ through the aggregated update.\footnote{$C_{FTTE}$ and $D_{FTTE}$ arise from the same two ingredients as in the FedBuff analysis: (i) bounding the deviation of the aggregated update from the true gradient direction, 
$\mathbb{E}[\Delta_t \mid \mathcal{F}_t] - \alpha(Q) \nabla f(w^t)$, and (ii) bounding its second moment $\mathbb{E}[\|\Delta_t\|^2]$, as detailed in the proofs (in Appendix.D of \cite{nguyen2022federatedlearningbufferedasynchronous}) leading to Eq. (2) and Eq. (4) in \cite{nguyen2022federatedlearningbufferedasynchronous}. In FedBuff these bounds are written for the unweighted average $\Delta_t = \frac{1}{K} \sum_{i \in S_t} \Delta_t^i$, whereas in FTTE we instead have a reweighted aggregate $\Delta_t^{\text{FTTE}} = \frac{1}{K} \sum_{i \in S_t} \omega_i(t) \Delta_t^i, \quad \text{with} \quad \omega_i(t) = \frac{1}{1 + \tau_i(t) v_i(t)}$, so the same arguments apply but the resulting constants, $C_{FTTE}$ and $D_{FTTE}$, now encode the effect of age–variance weighting on these bias and variance terms.} Intuitively, smaller $C_{\text{FTTE}}$ and $D_{\text{FTTE}}$ correspond to reduced noise and weaker impact of stale updates.

\begin{corollary}
Let us choose learning rates $\eta_l$
and $\eta_g$ such that $\eta_l \eta_g Q \leq \frac{1}{L}$, $\eta_l = \mathcal{O} \left( \frac{1}{K\sqrt{TQ}} \right)$,  $\eta_g = \mathcal{O}(K)$, $F^* := f(w^0) - f(\hat{w}^*)$, and sufficiently large $T$. Then we have
\begin{equation}
\begin{split}
    \frac{1}{T} \sum_{t=0}^{T-1} \mathbb{E}\| \nabla f(w^t) \|^2 \leq & \mathcal{O}(\frac{F^*}{\sqrt{TQ}}) + \mathcal{O}(\frac{C_{\text{FTTE}}}{\sqrt{TQ}} ) \\
    & + \mathcal{O}(\frac{D_{\text{FTTE}}}{T}) + \mathcal{O}(\Delta_{approx}) 
\end{split}
\end{equation}
\end{corollary}

Corollary 1 mirrors the FedBuff corollary; for an appropriate choice of constant learning rates, it yields the same non‑convex rate with FTTE‑specific constants and the constrained optimum $\hat{w^*}$. By Theorem 1 and Corollary 1, FTTE attains the same FedBuff‑style non‑convex convergence rate, namely an ergodic gradient norm of order $\mathcal{O}(\frac{1}{\sqrt{TQ}})$, plus an additive approximation term $\Delta_{approx}$. In particular, its worst‑case iteration and communication complexity are asymptotically no better and no worse than those of FedBuff. Under Assumption 6, however, clients that are both stale and high‑variance $(\tau_i(t) \geq 1,v_i(t) \geq 1)$,
satisfy 
$$\omega_i^{\text{FTTE}}(t) = \frac{1}{1 + \tau_i(t) v_i(t)} \leq \frac{1}{1 + \tau_i(t)} = \omega_i^{\text{age}}(t)$$
so their updates are down‑weighted more aggressively than under age‑only staleness. This reduces the influence of harmful updates in the variance and staleness terms that define the constants $C_{FTTE}$ and $D_{FTTE}$, which we therefore expect to be smaller in practice and which matches the substantially fewer communication rounds FTTE needs to reach a given target accuracy in our experiments (Section \ref{eval}).

\section{Experiments \& Results}
\label{eval}

\textbf{Implementation Details: } For parameter selection, pre-trained models are used and they are optimized for a memory budget, $M_{min}=64$ MB on TinyImageNet assuming that local data at the clients is private. The parameter-selection step involves evolutionary search, which is theoretically expensive. However in practice, this takes only a few minutes on a GPU and we assume that the server is resource-abundant. Additionally, this step is only performed once before the start of the federated training. The memory budget depends on the devices in the real-world network. Downstream datasets are - Oxford-IIIT Pet \cite{oxford-iiit}, CIFAR-10 \cite{krizhevsky2009learning}, Oxford Flowers \cite{Nilsback08}, and Skin Cancer diagnosis\cite{9121248}. To model data heterogeneity, training data is partitioned across clients using a Dirichlet distribution \cite{tenison2023gradient} with $\alpha \in \{100000, 0.1\}$, representing IID and Non-IID distributions respectively. Experiments use — MobileNetV2\cite{sandler2019mobilenetv2invertedresidualslinear}, MCUNet\cite{lin2020mcunettinydeeplearning}, and ProxylessNAS\cite{cai2019proxylessnasdirectneuralarchitecture} - models that are widely adopted in edge and IoT research. 

Unless otherwise mentioned, all experiments simulate a network of 100 clients with Dirichlet $\alpha=1.0$ representing a moderately heterogenous data distribution, and randomly chosen 50\% stragglers and delays of upto 30 seconds. Most experiments consider FedAVG or SyncFL as the baseline as other baselines often fail to achieve target accuracy of FTTE as shown in Table\ref{iid}. SGD with learning rate 0.1, 3 local epochs, and a batch size of 8 are used for local updates. FedBuff and FTTE use a buffer size of 10 as recommended in \cite{nguyen2022federatedlearningbufferedasynchronous}. Note that FTTE is modular and can be readily integrated with advanced techniques such as quantization for further resource savings or with techniques like differential privacy and secure aggregation for added privacy, or with recent state-of-the-art FL algorithms to leverage their benefits in heterogeneity, optimization stability, or robustness. Additionally, adaptive optimizers \cite{reddi2021adaptivefederatedoptimization} showed similar results as non-adaptive optimizers presented in this section. We also implement FTTE on a real resource constrained federated network of four Raspberry Pis and the results indicate its real-world deployability. The repository will be open-sourced on acceptance.

\begin{figure}[t]
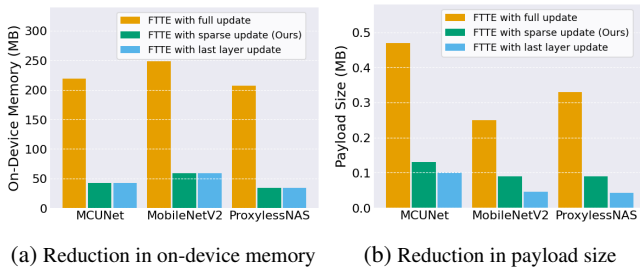

    \centering
    \begin{subfigure}[b]{0.239\textwidth}
        \includegraphics[width=\linewidth]{images/memory.pdf}
        
        \subcaption{\footnotesize Reduction in on-device memory}
        \label{mem}
    \end{subfigure}
    \hfill
    \begin{subfigure}[b]{0.239\textwidth}
        \includegraphics[width=\linewidth]{images/payload.pdf}
        
        \subcaption{\footnotesize Reduction in payload size}
        \label{pay}
    \end{subfigure}
        \caption{(a) On-device memory and (b) payload size requirements for FTTE with full updates as in classic FL, last layer update as in TL, and sparse update (ours). }
    \label{TL}
\end{figure}

\subsection{Faster Convergence} \label{table} FTTE achieves markedly faster convergence than all the considered FL baselines, as quantified by the number of required communication steps (one round equals a model upload/download). This rapid convergence enables earlier deployment of performing models, significantly reduces overall training time, and thus enhances device energy efficiency and lifespan by limiting both runtime and thermal load. Table \ref{iid} compares FTTE with SyncFL ie, FedAVG \cite{mcmahan2023communicationefficientlearningdeepnetworks}, AsyncFL \cite{xie2020asynchronousfederatedoptimization}, and Semi-AsyncFL ie, FedBuff \cite{nguyen2022federatedlearningbufferedasynchronous}, across a range of models, datasets, and data heterogeneity. In each setting, FTTE is trained until it meets the target accuracy, while baselines are evaluated by the number of steps needed to reach the same target accuracy. If a method surpasses the target accuracy (e.g., FedAVG), we log the round when the threshold is first met. When a baseline does not attain the target accuracy within 10k communicationsteps or converges to an accuracy lower than target accuracy (often observed for FedBuff), the number of rounds is recorded as “$>$10k.” Unstable runs resulting in oscillating loss are marked as “Osc.” — a failure mode prevalent for AsyncFL and non-IID baselines. As evidenced by Table\ref{iid} and Fig.\ref{comm_main}, \textit{\textbf{FTTE consistently matches the target accuracy in far fewer communication steps across all evaluated scenarios.}} including IID and Non-IID data distributions simulated using Dirichlet $\alpha=10000$ and $\alpha=0.1$ respectively.

\begin{figure}[t]
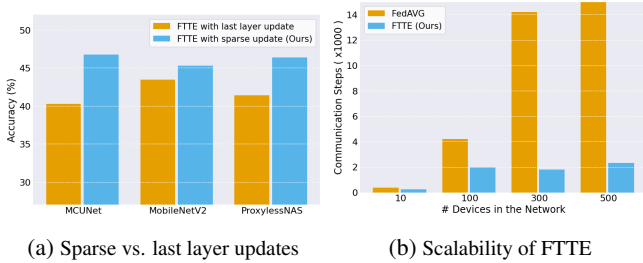

    \centering
    \begin{subfigure}[b]{0.239\textwidth}
        \includegraphics[width=\linewidth]{images/tl_sp_acc.pdf}
        \subcaption{\footnotesize Sparse vs. last layer updates}
        \label{tlvssp}
    \end{subfigure}
    \hfill
    \begin{subfigure}[b]{0.239\textwidth}
        \includegraphics[width=\linewidth]{images/scalability.pdf}
        
        \subcaption{\footnotesize Scalability of FTTE}
        \label{scalablity}
    \end{subfigure}
        \caption{(a) Convergence accuracy of sparse updates versus last layer update scheme. (b) FTTE is a scalable system with upto 500 devices with randomly chosen 50\% stragglers.}

\end{figure}

\subsection{Memory and Payload Efficiency}
FTTE achieves substantial improvements in both memory utilization and payload efficiency. As demonstrated in Fig.\ref{main}, FTTE reduces on-device memory consumption for local training by 80\% and communication payload by 69\% compared to FL methods that employ full model update strategies on CIFAR-10. This efficiency arises directly from FTTE’s sparse model update mechanism (see Subsec.\ref{Fl}). As shown in Fig.\ref{TL} showing results on Skin Cancer dataset, FTTE’s sparse update scheme consistently maintains client memory usage below the threshold ($M_\mathrm{min}=64$MB) and yields markedly lower memory and transmission payloads than full update baselines. While limiting updates to only the last layer (as in typical TL) achieves slightly lower memory and payload, this approach incurs up to a 7\% drop in accuracy versus FTTE’s sparse selection (see Fig.\ref{tlvssp}), confirming that \textit{\textbf{FTTE offers the best trade-off between efficiency and model performance for resource-constrained FL}}. Additionally, while FTTE currently applies sparse updates without additional compression or hardware co-design, it is orthogonal and complementary to techniques such as quantization or activation scaling, offering further avenues for improving memory efficiency.

\begin{figure}[t]
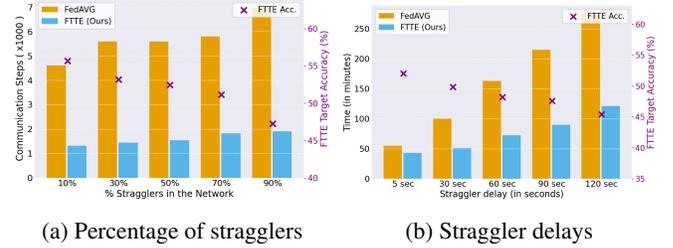

    \centering
    \begin{subfigure}[b]{0.239\textwidth}
        \includegraphics[width=\linewidth]{images/stragglers_with_accuracy.pdf}
        
        \subcaption{Percentage of stragglers}
        \label{straggler_percent}
    \end{subfigure}
    \hfill
    \begin{subfigure}[b]{0.239\textwidth}
        \includegraphics[width=\linewidth]{images/stragglers_2_with_acc.pdf}
        
        \subcaption{Straggler delays}
        \label{delay}
    \end{subfigure}
    
    \label{straggler_delay}
    
        \caption{ Shows communication steps required for  FedAVG and FTTE under (a) increasing percentage of stragglers and (b) increasing delay per straggling client in the federated network. }
\end{figure}

\subsection{Effect of Stragglers}
FTTE is explicitly designed to ensure robust FL in networks heavily impacted by stragglers, particularly those comprised of memory-limited edge devices. As illustrated in Fig.\ref{straggler_percent} and Fig.\ref{delay}, FTTE consistently outperforms FedAVG or SyncFL as the proportion of stragglers increases or as individual device delays grow severe. When straggler rates or percent of straggling devices in the federated network reach 90\% or client delay of 50\% stragglers vary from 5 to 120 seconds, FedAVG’s communication steps  and total wall-clock training time escalate sharply - up to nearly 7,000 steps and 275 minutes - respectively, while FTTE incurs only a modest increase. Notably, FTTE maintains competitive model accuracy despite high straggler rates or delay, exhibiting only minimal degradation in these adverse settings as shown in Fig.\ref{straggler_percent} and Fig.\ref{delay}. \textit{\textbf{These results demonstrate that FTTE effectively mitigates straggler-induced inefficiency, providing both high efficiency and robust learning performance under real-world conditions}}.
\begin{table}[]
\caption{Target accuracy and communication steps required by FTTE with Adam optimizer (as local optimizer) and FedAdam \cite{reddi2021adaptivefederatedoptimization} to achieve target accuracy on MCUNet with different datasets and data distributions. This Federated network consists of 100 clients with 50\% stragglers.}
\label{adaptive}
\resizebox{\columnwidth}{!}{%
\begin{tabular}{ccccc}
\hline
\multicolumn{5}{c}{\cellcolor[HTML]{EFEFEF}\textbf{MCUNet}}                                                                                                                                                                                                                                                                                                                           \\ \hline
\multicolumn{1}{l}{}                                                                          &                                                    &                                                                                  &                                                            &                                                                                            \\
\multicolumn{1}{l}{\multirow{-2}{*}{}}                                                        & \multirow{-2}{*}{\textbf{$\alpha$}}                & \multirow{-2}{*}{\textbf{Target Acc.}} & \multirow{-2}{*}{\textbf{FedAdam}}                         & \multirow{-2}{*}{\textbf{\begin{tabular}[c]{@{}c@{}}FTTE (Ours)\\ with Adam\end{tabular}}} \\ \hline
                                                                                              & \cellcolor[HTML]{F5F5F5}100000                     & \cellcolor[HTML]{F5F5F5}45.67\%                                                    & \cellcolor[HTML]{F5F5F5}6200 ($\times$7.6)                        & \cellcolor[HTML]{F5F5F5}815                                                                \\
                                                                                              & 1.0                                                & 43.70\%                                                                            & 4400 ($\times$3.5)                                                & 1269                                                                                       \\
\multirow{-3}{*}{\textbf{\begin{tabular}[c]{@{}c@{}}Skin Cancer\\ (8 classes)\end{tabular}}}  & \cellcolor[HTML]{F5F5F5}{\color[HTML]{000000} 0.1} & \cellcolor[HTML]{F5F5F5}{\color[HTML]{333333} 30.05\%}                             & \cellcolor[HTML]{F5F5F5}{\color[HTML]{333333} 2600 ($\times$2.7)} & \cellcolor[HTML]{F5F5F5}{\color[HTML]{333333} 969}                                         \\ \hline
                                                                                              & 100000                                             & 67.46\%                                                                            & 9200 ($\times$7.6)                                                & 1211                                                                                       \\
                                                                                              & \cellcolor[HTML]{F5F5F5}1.0                        & \cellcolor[HTML]{F5F5F5}63.57\%                                                    & \cellcolor[HTML]{F5F5F5}8000 ($\times$6.7)                        & \cellcolor[HTML]{F5F5F5}1300                                                               \\
\multirow{-3}{*}{\textbf{CIFAR-10}}                                                           & 0.1                                                & 50.70\%                                                                            & 6600 ($\times$5.2)                                                & 1267                                                                                       \\ \hline
                                                                                              & \cellcolor[HTML]{F5F5F5}100000                     & \cellcolor[HTML]{F5F5F5}67.34\%                                                    & \cellcolor[HTML]{F5F5F5}4800 ($\times$4.2)                        & \cellcolor[HTML]{F5F5F5}1135                                                               \\
                                                                                              & 1.0                                                & 66.02\%                                                                            & 3665 ($\times$4.5)                                                & 815                                                                                        \\
\multirow{-3}{*}{\textbf{\begin{tabular}[c]{@{}c@{}}Oxford-PETS\\ (37 classes)\end{tabular}}} & \cellcolor[HTML]{F5F5F5}0.1                        & \cellcolor[HTML]{F5F5F5}56.02\%                                                    & \cellcolor[HTML]{F5F5F5}2900 ($\times$1.5)                        & \cellcolor[HTML]{F5F5F5}1915                                                               \\ \hline
                                                                                              & 100000                                             & 67.28\%                                                                            & 5400 ($\times$5.3)                                                & 1017                                                                                       \\
                                                                                              & \cellcolor[HTML]{F5F5F5}1.0                        & \cellcolor[HTML]{F5F5F5}65.69\%                                                    & \cellcolor[HTML]{F5F5F5}5000 ($\times$4.5)                        & \cellcolor[HTML]{F5F5F5}1101                                                               \\
\multirow{-3}{*}{\textbf{Flowers-102}}                                                        & 0.1                                                & 64.56\%                                                                            & 6900 ($\times$3.5)                                                & 1981                                                                                       \\ \hline
\end{tabular}
}
\end{table}
\subsection{Scalability} 
FTTE exhibits strong scalability as the network size increases, maintaining efficient convergence with significantly fewer communication steps compared to baseline methods (Fig.\ref{scalablity}). While FedAVG’s required communication steps exceed 15,000 when scaling to 500 devices (our experiments are limited to 500 clients due to compute constraints), FTTE displays only a mild growth, underscoring its communication efficiency for large-scale FL. Notably, FTTE is approximately $7.5\times$ faster than FedAVG or SyncFL on a 500-device network, \textit{\textbf{confirming its suitability for deployment in extensive, heterogeneous edge environments.}} 

\subsection{Effect of Variance on Staleness Function} 
Table \ref{staleness} compares the proposed age–variance staleness function, $\frac{1}{1+Age*Var}$, against conventional age-only weighting schemes $\frac{1}{1+Age}$ and $\frac{1}{(1+Age)^2}$. Across both Oxford Pets and CIFAR-10, the age–variance formulation consistently achieves higher target accuracy with fewer communication steps. On Oxford Pets, it reaches 58.3\% accuracy in 925 steps, outperforming $\frac{1}{1+Age}$ (1157 steps) and $\frac{1}{(1+Age)^2}$ (53.65\% in 995 steps). Similarly, on CIFAR-10, it attains 72.21\% accuracy in 771 steps, improving both convergence speed and final accuracy. These results demonstrate that \textit{\textbf{incorporating update variance alongside temporal staleness enables more effective discrimination of client updates}}, leading to an empirically observed 
$\approx 20\% $ improvement in convergence efficiency and accuracy over age-only schemes.

\subsection{Extension to other federated optimizers}
Table~\ref{adaptive} compares FTTE combined with the Adam optimizer against FedAdam  \cite{reddi2021adaptivefederatedoptimization} across multiple datasets and levels of data heterogeneity. Across all evaluated datasets—Skin Cancer, CIFAR-10, Oxford-PETS, and Flowers-102—FTTE consistently reaches the target accuracy in substantially fewer communication rounds. On the Skin Cancer dataset with near-IID data $\alpha=10000$, FTTE converges in 815 steps compared to 6200 for FedAdam (7.6× reduction). Under varying heterogeneity $\alpha=\{10000, 1, 0.1\}$, FTTE preserves a 
2.7 - 3.5× communication efficiency advantage while maintaining comparable or improved accuracy. These results indicate that \textit{\textbf{FTTE is complementary to federated optimizers and algorithms}} and remains effective across a wide range of data distributions.

\begin{table}[t]
    \centering
    \caption{Target accuracy and communication steps to reach target accuracy of FTTE with different staleness functions including the proposed age-variance function.}
    \label{staleness}
\begin{tabular}{clcc}
\hline
                                                                        &                                                 & \textbf{\begin{tabular}[c]{@{}c@{}}Target\\ Acc.\end{tabular}} & \textbf{\begin{tabular}[c]{@{}c@{}}Comm.\\ Steps\end{tabular}} \\ \hline
                                                                        & \cellcolor[HTML]{EFEFEF}$1/(1+age)$             & \cellcolor[HTML]{EFEFEF}58.42 \%                               & \cellcolor[HTML]{EFEFEF}1157                                   \\
                                                                        & $1/(1+age)^2$                                   & 53.65 \%                                                       & 995                                                            \\
\multirow{-3}{*}{\begin{tabular}[c]{@{}c@{}}Oxford\\ Pets\end{tabular}} & \cellcolor[HTML]{EFEFEF}$1/(1+age*Var.)$   & \cellcolor[HTML]{EFEFEF}58.30 \%                                   & \cellcolor[HTML]{EFEFEF}925                                    \\ \hline
                                                                        & \cellcolor[HTML]{EFEFEF}$1/(1+age)$             & \cellcolor[HTML]{EFEFEF}70.50 \%                                & \cellcolor[HTML]{EFEFEF}881                                    \\
                                                                        & $1/(1+age)^2$                                   & 68.77 \%                                                       & 897                                                            \\
\multirow{-3}{*}{\begin{tabular}[c]{@{}c@{}}CIFAR\\ 10\end{tabular}}    & \cellcolor[HTML]{EFEFEF}$1/(1+age*Var.)$   & \cellcolor[HTML]{EFEFEF}72.21 \%                               & \cellcolor[HTML]{EFEFEF}771                                    \\ \hline
\end{tabular}
\end{table}

\begin{table}[]
\centering
\caption{Target accuracy and communication steps required to reach the accuracy by FTTE on a federated network of 4 Raspberry Pis. }
\label{tab:ondevice-rpi}
\begin{tabular}{ccc}
\hline
                      & \textbf{Target Acc. (\%)} & \textbf{Comm. Steps} \\ \hline
\rowcolor[HTML]{EFEFEF} 
\textbf{MCUNet}       & 72.27                     & 19                 \\
\textbf{MobileNetV2}  &        70.32            &           19        \\
\rowcolor[HTML]{EFEFEF} 
\textbf{ProxylessNAS} &       71.25                      &          13           \\ \hline
\end{tabular}
\end{table}

\subsection{On-Device Experiments}
We evaluate FTTE in a real on-device setup using a federated network of four Raspberry Pi 5 devices (16 GB RAM each) training CIFAR-10 with Dirichlet distribution $\alpha=1.0$. Each Pi participates as a client running local FTTE updates on MCUNet, MobileNetV2, or ProxylessNAS. Table~\ref{tab:ondevice-rpi} reports the target test accuracy and the number of communication steps required by FTTE to reach that accuracy; FTTE attains 72.27\% with MCUNet in only 19 rounds, 
70.32\% with MobileNetV2 in 19 rounds, and 71.25\% with ProxylessNAS in 13 rounds, demonstrating that the method remains communication-efficient and stable even on modest edge hardware. 

\subsection{Future Work}
FTTE opens several promising directions. First, our current analysis assumes a fixed sparse subset and fixed staleness weights. Extending the theory and system to adaptive parameter selection and dynamic reweighting is an important next step that would help with systems that are dynamic. Second, while we evaluated moderate-scale image models, applying FTTE to larger foundation models and sequence tasks (e.g., language, time series, audio) would test its limits under more extreme memory and bandwidth constraints. Third, integrating FTTE with complementary techniques such as quantization, compression, and formal privacy guarantees (secure aggregation, differential privacy) could further reduce resource usage while preserving robustness in heterogeneous, real-world deployments.

\section{Conclusion}
We introduce FTTE, a novel FL framework designed for federated edge intelligence networks with devices having severe memory and communication limits. FTTE uniquely integrates memory-aware parameter selection with sparse updates and communication and age- and variance-weighted aggregation, enabling robust and sparse semi-asynchronous training under extreme heterogeneity and straggling. Extensive experiments demonstrate that FTTE not only delivers 81\% faster convergence, 80\% memory and 69\% communication payload reduction on CIFAR-10 and other datasets, but also consistently achieves higher accuracy than FedBuff, including regimes with up to 500 clients and 90\% stragglers. Similar results were observed on adaptive federated optimizers as well indicating the modular applicability of FTTE to a variety of federated algorithms. These results establish FTTE as a practical, scalable solution for real-world FL on heterogeneous, highly resource-constrained networks, advancing state-of-the-art in the intersection of robustness and resource efficiency.

\section{Acknowledgments}
The authors acknowledge that OpenAI’s ChatGPT was used to assist with the refinement and editing of the manuscript text. The content and technical contributions of this paper remain the sole responsibility of the authors.


\bibliographystyle{IEEEtran}
\bibliography{ref}


\end{document}